\pdfoutput=1

\documentclass[11pt]{article}

\usepackage{naacl2021}

\usepackage{times}
\usepackage{latexsym}

\usepackage{booktabs}
\usepackage{graphicx}
\usepackage{subcaption}
\usepackage{multirow}
\usepackage[hyphens]{xurl}
\usepackage{comment}

\usepackage[T1]{fontenc}

\usepackage[utf8]{inputenc}

\usepackage{microtype}

\title{Understanding Transformers for Bot Detection in Twitter}

\author{Andres Garcia-Silva \\
  Expert.ai Research Lab \\
  Prof. Waksman 10\\
  28036 Madrid, Spain \\
    \texttt{agarcia@expert.ai} \\\And
  Cristian Berrio \\
  Expert.ai Research Lab \\
  Prof. Waksman 10\\
  28036 Madrid, Spain \\
    \texttt{cberrio@expert.ai} \\\And
  Jose Manuel Gomez-Perez \\
  Expert.ai Research Lab \\
  Prof. Waksman 10\\
  28036 Madrid, Spain \\
    \texttt{jmgomez@expert.ai} \\
    
    }

\begin{document}
\maketitle
\begin{abstract}
In this paper we shed light on the impact of fine-tuning over social media data in the internal representations of neural language models. We focus on bot detection in Twitter, a key task to mitigate and counteract the automatic spreading of disinformation and bias in social media. We investigate the use of pre-trained language models to tackle the detection of tweets generated by a bot or a human account based exclusively on its content. Unlike the general trend in benchmarks like GLUE, where BERT generally outperforms generative transformers like GPT and GPT-2 for most classification tasks on regular text, we observe that fine-tuning generative transformers on a bot detection task produces higher accuracies. We analyze the architectural components of each transformer and study the effect of fine-tuning on their hidden states and output representations. Among our findings, we show that part of the syntactical information and distributional properties captured by BERT during pre-training is lost upon fine-tuning while the generative pretraining approach manage to preserve these properties.

\end{abstract}

\section{Introduction}

Nowadays fine-tuning neural networks trained on a language modeling objective \cite{Bengio:2003:NPL:944919.944966} has become the usual way to address most NLP tasks. 
While early language models like ELMo \cite{PetersELMo2018} and ULMFit \cite{howard2018ulmfit} used LSTM architectures, as of today transformers \cite{Vaswani2017Transformers} are the foundation of successful language models like GPT \cite{radford2018OpenAITransformer} and BERT \cite{devlin2018BERT}.

Most language models are trained on high quality, grammatically correct, curated text corpora such as  Wikipedia (ULMFiT), BookCorpus (GPT), a collection of Web Pages (GPT-2), a combination of Wikipedia and BookCorpus (BERT) or News (ELMo). In fact, BERT requires sequences of sentences that makes sense due to the additional next sentence prediction learning objective. 
However, none of these corpora cover text from social media where users, human or automated agents, are continuously sharing pieces of information as part of the social interaction.  This brings up the question  whether such language models also capture the nuances of the short, informal and evolving language often found in social media channels. 

In this paper we shed light on the impact of fine-tuning language models over social media data. Particularly, we focus on bot detection in Twitter, which is a key task to mitigate and counteract the automatic spreading of disinformation and bias in social media \cite{shao2018spread}. We model this task as a binary classification problem where the goal is to detect whether the content of a tweet has been produced by a bot or a human. Note that different features, e.g. number of followees and followers, number of retweets, presence of hashtags, URLs, etc., can be valuable sources of relevant information to solve this task.  However, it is not our objective to beat the SotA in bot detection, but to understand how pre-trained language models adapt their internal representations in the presence of such data and make an informed interpretation of the resulting predictions. Therefore, we focus solely on the analysis of the actual text contained in the tweet, which has been regarded as highly relevant to detect automatic accounts \cite{Gilani:2017:CTA:3110025.3110091}.

Looking at multi-task benchmarks like GLUE~\cite{wang-etal-2018-glue}, most tasks are dominated by transformer-based models fine-tuned on task-specific data. This is also the case for classification tasks, where a deeper look reveals that BERT-based models systematically outperform models based on generative pre-training like GPT. However, our experiments in a bot detection dataset (section~\ref{sec:data}) show that this trend actually reverses (section~\ref{sec:fine}) for the bot detection task. We investigate the reasons of this apparent anomaly by carrying out a bottom-up analysis on the transformers used in each language model, from vocabulary and word embeddings (section~\ref{sec:voc}) through hidden states to output representations (section~\ref{sec:hid}). Our results show that fine-tuning for bot detection impacts BERT and GPT-2 differently with respect to the contextualized outputs and linguistic properties learned in pre-training.


We observe that BERT's hidden states capture less grammatical information after fine-tuning, while GPT-2 manages to generate representations that preserve such information as in the pre-trained model. We show evidence that after fine-tuning GPT-2 tends to generate more distributed outputs and some clusters can be  roughly mapped to twitter entities while BERT outputs do not express this differentiation and are all concentrated in the same region of the vector space. For reproducibility purposes, the paper includes a GitHub repository\footnote{\url{https://github.com/botonobot/Understanding-Transformers-for-Bot-Detection-Twitter}} with experimental data and code, as well as detailed diagrams and figures.

\section{Related work}
\label{sec:rel}

Different neural architectures have been proposed to pre-train neural language models on large text corpus  \cite{howard2018ulmfit,radford2018OpenAITransformer,devlin2018BERT}
. Transformer architectures \cite{Vaswani2017Transformers} are used in GPT \cite{radford2018OpenAITransformer},  and BERT \cite{devlin2018BERT}, two language models that have pushed the state of the art in different NLP tasks, inspiring subsequent work like GPT-2 \cite{radford2019language}, RoBERTa \cite{liu2019roberta} and DistilBERT \cite{sanh2019distilbert}. 

Recently, language models are being used to process text from social media. BERT is used for the classification of tweets in the disaster management field \cite{Ma2019TweetsCW}, hate speech identification \cite{Mishra20193IdiotsAH,Mozafari2019ABT}, offensive language classification \cite{Paraschiv2019UPBAG}, subjectivity, polarity and irony detection \cite{polignano2019alberto}, and to evaluate tweet predictability~\cite{Przybya2019DetectingBA}. The generative ability of GPT-2 is  used to replicate politician's twitter accounts \cite{Ressmeyer2019DeepFP}, and to generate fake news and enable its detection \cite{zellers2019defending}. 

Since pre-trained language models work very well in practice, 
researchers are trying to understand how they work \cite{rogers2020primer}.  \newcite{peters-etal-2018-dissecting} found that language models encode a rich  hierarchy of contextual information throughout the layers, from morphological at the embeddings, local-syntax at lower layers and semantic information at upper layers. \newcite{tenney-etal-2019-bert} show that 
BERT encodes phrase-level information in its lower layers and a hierarchy of linguistic information in its intermediate layers. \newcite{ethayarajh2019contextual} shows that BERT and GPT-2 hidden states and outputs occupy a narrow
cone in the vector space. \newcite{clark-etal-2019-bert} investigated self attention and found some heads that capture syntactical information. \newcite{kovaleva-etal-2019-revealing} shows that in fine-tuning the attention mechanism in BERT focus mainly on the \textit{CLS} and \textit{SEP} tokens. Our research work can be framed with the above mentioned since our goal is to understand the effect of fine-tuning on the hidden states and outputs of the language models by put them into test on tasks related to bot detection, grammatical analysis and contextual similarity, and provide reference data from classification tasks in standard benchmarks.



Automatic accounts known as bots publish a large amount of information in social media, and some of them take active part on spreading misinformation and fake news \cite{shao2018spread}. In fact, twitter points in its last report about platform manipulation\footnote{\url{https://transparency.twitter.com/en/reports/platform-manipulation.html}} at malicious bots as one of the main actors, along with spam, and fake accounts. 

In general, detecting bots can be addressed as a binary classification problem that can leverage different features from the social network including user metadata, user relations, and posting activity, as well as the content of the posts. Approaches like Botomoter \cite{Varol2017OnlineHI} work at the account level and use all the available information to derive features that are processed with an ensemble of machine learning algorithms. However this approach shows a large disagreement with human annotators \cite{Gilani:2017:CTA:3110025.3110091}, and leads to false positives and negatives \cite{rauchfleisch2020false}. In this paper we work at the tweet-level focusing only on the tweet content to determine whether it was posted by a bot or a human. In \cite{Gilani:2017:CTA:3110025.3110091} human annotators cited the style and the patterns of the tweets as strong indicators of automatic accounts.

\section{Dataset}
\label{sec:data}

We tap into a ground truth dataset of bot and human accounts released by~\newcite{Gilani:2017:CTA:3110025.3110091}. This dataset was manually annotated by 4 annotators, achieving 89\% inter-annotator agreement, 77  Cohen’s kappa ($\kappa$) coefficient indicating substantial agreement. We generate a balanced dataset containing 600,000 tweets labelled as bot or human according to the account label. Our training dataset comprises a total of 500,000 tweets where 279,495 tweets were created by 1,208 human accounts, and 220,505 tweets were tweeted from 722 bot accounts. The test set contains the remaining 100,000 tweets where 55,712 are tweets from human accounts and 44,288 are tweets from bot accounts. 

Different usage patterns emerge in this dataset for bots and humans regarding twitter entities such as user mentions, hashtags and URL. For example,  bots are more active averaging 305 tweets per account compared to 231 tweets of humans. Bots also seek to redirect more traffic and increase visibility using 0.8313 URL and 0.4745 hashtags per tweet, while humans use 0.5781 URL and 0.2887 hashtags. Finally, humans interact more with other accounts by mentioning other users 0.5781 times per tweet, while bots display a less social behaviour mentioning other users only 0.4371 times per tweet.

\section{Fine-tuning transformer language models for bot detection}
\label{sec:fine}

We follow~\newcite{devlin2018BERT} and~\newcite{radford2018OpenAITransformer} to tackle single sequence classification with BERT and GPT respectively. That is, we add a linear layer on top of the last hidden state of the classification token \textit{CLS} (BERT) or the last token (GPT) to train a softmax classifier. The only new parameters to train from scratch are the weights in the new layer $W \in \mathbf{R}^{KxH}$, where $k$ is the number of classification labels and $H$ is the size of the hidden state representation for \textit{CLS} or the last token. In our experiments we use BERT base and GPT, that are comparable models in size (110M parameters), and the GPT-2 version with 117M parameters. These models are available in the Transformers library~\cite{Wolf2019HuggingFacesTS}. 
We use batch size 8 and train during 2 epochs. As the f-measure values indicate, the evaluation results reported in table \ref{tab:lmbotdetection} show that GPT and GPT-2 consistently achieved better performance in this task (henceforth we focus on GPT-2 rather than GPT since the former generally outperforms the latter in bot detection). 

To evaluate the resulting models we designed a preliminary experiment to spot twitter entities like hashtags, user mentions, and URLs. Spotting twitter entities is relevant for bot detection since bots and humans seem to use them in different ways, e.g. bots tend to focus on hashtags rather than users, while the opposite applies for humans. We fine-tune the language models following the same procedure used by \newcite{devlin2018BERT} to fine-tune BERT on SQuAD v1.1 \cite{rajpurkar-etal-2016-squad}. We use an entity type as a question and the tweet content as paragraph so that the task is to identify the entity text span in the tweet. For instance, to supervise BERT on this task we use sequences such as  "\textit{[CLS] hashtag? [SEP] Get ready for the next \#YouthOlympics [SEP]}", and the expected answer is "\textit{\#YouthOlympics}". 

\begin{table}[tb!]
  \centering
    \footnotesize
    \begin{tabular}{llrrr}
    \toprule
    Model & Library & \multicolumn{1}{c}{Precision} & \multicolumn{1}{c}{Recall} & \multicolumn{1}{c}{F-score} \\
    \midrule
    GPT-2 & OpenAI & 0.8657 & 0.8640 & \textbf{0.8630} \\
    GPT   &  OpenAI & 0.8567 & 0.8546 & \textbf{0.8533} \\
    GPT-2 & Transformers & 0.8549 & 0.8524 & \textbf{0.8509} \\
    BERT  & Transformers & 0.8443 & 0.8437 & 0.8439 \\
    BERT  & Google-bert & 0.8572 & 0.8213 & 0.8388 \\
    SVM  & LinearSVC & 0.8033 & 0.8031 & 0.8032 \\
    \bottomrule
    \end{tabular}%
  \caption{Fine-tuned language models on the bot detection task. We trained a linear SVM baseline \protect\cite{Fan2008LibLinear} with a tuned regularization parameter, lemmatized words using wordnet, and stopwords removed.}
  \label{tab:lmbotdetection}%
\end{table}%


\begin{table*}[htbp]
  \centering
  \footnotesize
    \begin{tabular}{clllrrrrr}
          &       & \multicolumn{3}{c}{\textbf{Entity anywhere}} &       & \multicolumn{3}{c}{\textbf{Entity at the end}} \\
\cmidrule{3-5}\cmidrule{7-9}    \textbf{Twitter Entity} & \multicolumn{1}{c}{\textbf{Model}} & \multicolumn{1}{c}{\textbf{start}} & \multicolumn{1}{c}{\textbf{end}} & \multicolumn{1}{c}{\textbf{span}} &       & \multicolumn{1}{c}{\textbf{start}} & \multicolumn{1}{c}{\textbf{end}} & \multicolumn{1}{c}{\textbf{span}} \\
    \midrule
    \multirow{2}[0]{*}{Hashtag} & GPT-2  & \textbf{0.6673} & \textbf{0.2351} & \textbf{0.2582} &       & \textbf{0.8618} & \textbf{0.9946} & \textbf{0.9150} \\
          & BERT  & 0.3625 & 0.1596 & 0.1011 &       & 0.4649 & 0.5513 & 0.2452 \\
    \multirow{2}[0]{*}{URL} & GPT-2  & \textbf{0.8649} & \textbf{0.8427} & \textbf{0.8112} &       & \textbf{0.9882} & \textbf{0.9984} & \textbf{0.9914} \\
          & BERT  & 0.7975 & 0.4927 & 0.5004 &       & 0.9828 & 0.6412 & 0.6923 \\
    \multirow{2}[1]{*}{User mention} & GPT-2  & \textbf{0.8242} & \textbf{0.5120} & \textbf{0.5047} &       & \textbf{0.7349} & \textbf{0.9954} & \textbf{0.8123} \\
          & BERT  & 0.7008 & 0.4066 & 0.3731 &       & 0.1583 & 0.7538 & 0.1359 \\          
    \end{tabular}%
    \caption{F-measure of the fine-tuned language models to spot twitter entities.}
  \label{tab:entityspotting}%
\end{table*}%

The evaluation results of the twitter entity spotting task are shown in table \ref{tab:entityspotting}. Note that we discriminate between entities anywhere and the ones appearing only at the end of the tweet in an attempt to find out whether removing the context on the right of the entities could have a positive effect. Regardless of the entity placement, GPT-2 outperforms BERT. For entities anywhere, GPT-2 learns to detect the start of all entities types, while for the end position it only learns to detect URLs. BERT predictions are less accurate for both start and end tokens in all entity types, and in fact it does not learn to detect the end token of any entity type. 


At this stage it is hard to distinguish what makes one language model more accurate spotting entities than the other beyond the fitness of each model's output for the task. We expected that removing the context on the right would largely improve BERT's results, considering that the right-to-left in the bidirectional and simultaneous text processing was part of the problem to generate good representations for this task, or in general to process tweets. Although we showed that removing context on the right benefited BERT in some cases, GPT-2 seems to benefit the most of this setting.

\section{Vocabulary and word embeddings}
\label{sec:voc}

To deal with rare words language models use fixed-size vocabularies of sub-word units. Sub-word units are a flexible representation between characters and word level models that allow representing infrequent words by putting together sub-word units. The vocabulary size of BERT base is 30,000 and is based on the WordPiece model \cite{Wu2016GooglesNM} while Open AI GPT-2 uses Byte Pair Encoding \cite{sennrich-etal-2016-neural}, with a vocabulary size of 50,257. For example, the tweet "\textit{Congratulations @LeoDiCaprio \#Oscars https://t.co/5WLesgfnbe}"\footnote{tweet from OMGtrolls : https://twitter.com/OMGtrolls} is converted to the sequences of word and sub-word units in table \ref{tab:sub-words}.  

\begin{table*}[htbp]
  \centering
    \resizebox{\textwidth}{!}{%
    \begin{tabular}{c|llllllllllll}
    \textbf{Model} &  \multicolumn{12}{c}{\textbf{Words and sub-words}} \\
    \hline
    \multirow{2}[2]{*}{BERT} & 'congratulations' & '@'   & 'leo' & '\#\#dic' & '\#\#ap' & '\#\#rio' & '\#'  & 'oscar' & '\#\#s' & 'https' & \multicolumn{1}{l}{':'} & \multicolumn{1}{l}{'/'} \\
          & /'    & 't'   & '.'   & 'co'  & '/'   & '5'   & '\#\#wl' & '\#\#es' & '\#\#gf' & '\#\#nb' & \multicolumn{1}{l}{'\#\#e'} &  \\
    \hline
    \multirow{2}[2]{*}{GPT-2} & 'cong' & 'ratulations' & 'Ġ@'  & 'le'  & 'odic' & 'ap'  & 'rio' & 'Ġ\#' & 'osc' & 'ars' & \multicolumn{1}{l}{'Ġhttps'} & \multicolumn{1}{l}{'://'} \\
          & 't'   & '.'   & 'co'  & '/'   & '5'   & 'w'   & 'les' & 'g'   & 'fn'  & 'be'  &       &  \\
    \hline
    \end{tabular}%
    }
    \caption{Word and sub-word units produced by BERT and GPT-2 tokenizations for the tweet "\textit{Congratulations @LeoDiCaprio \#Oscars https://t.co/5WLesgfnbe"}}
  \label{tab:sub-words}%
\end{table*}%

\begin{figure*}[htbp]
    \centering
    \begin{subfigure}[t]{.25\textwidth}
        \centering
        \includegraphics[width=\textwidth]{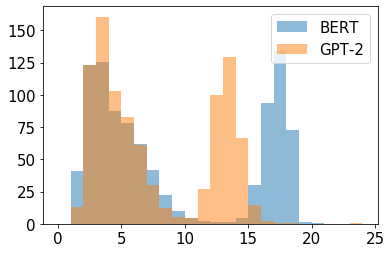}
        \caption{Bot detection}
        \label{fig:sub1}
    \end{subfigure}%
    \begin{subfigure}[t]{.25\textwidth}
        \centering
        \includegraphics[width=\textwidth]{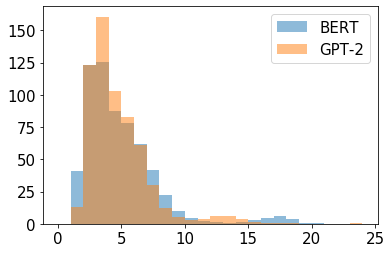}
        \caption{Bot detection no URLs}
        \label{fig:sub2}
    \end{subfigure}
    \begin{subfigure}[t]{.24\textwidth}
        \centering
        \includegraphics[width=\textwidth]{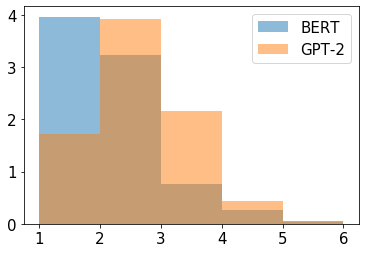}
        \caption{CoLA}
        \label{fig:sub3}
    \end{subfigure}
    \begin{subfigure}[t]{.24\textwidth}
        \centering
        \includegraphics[width=\textwidth]{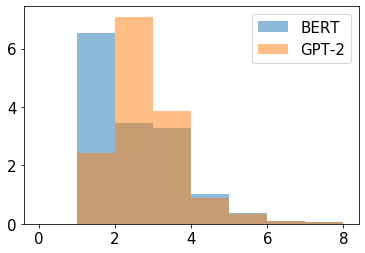}
        \caption{SST-2}
        \label{fig:sub4}
    \end{subfigure}\\ \bigskip
    \caption{Word length distribution on the bot detection, CoLA, and SST-2 datasets. X-axis is number of tokens in a word, and Y-axis is number of words in thousands}
    \label{fig:tok2enlenghtdist2}
  \end{figure*}

\begin{table}[htbp!]
  \centering
  \footnotesize
    \scalebox{0.9}{
    \begin{tabular}{rlrrr}
    \toprule
    \multicolumn{1}{l}{Dataset} & Model & \multicolumn{1}{l}{Precision} & \multicolumn{1}{l}{Recall  } & \multicolumn{1}{l}{F-score} \\
    \midrule
    \multicolumn{1}{l}{Bot detection} & GPT-2 & 0.7098 & 0.7120 & 0.7061 \\
          & BERT  & 0.7298 & 0.7274 & \textbf{0.7283} \\
    \multicolumn{1}{l}{CoLA} & GPT-2 & 0.4779 & 0.6913 & \textbf{0.5651} \\
          & BERT  & 0.6394 & 0.5139 & 0.5254 \\
    \multicolumn{1}{l}{SST-2} & GPT-2 & 0.5047 & 0.5092 & 0.3496 \\
          & BERT  & 0.6908 & 0.6319 & \textbf{0.6048} \\
    \multicolumn{1}{l}{MRPC} & GPT-2 & 0.7873 & 0.6912 & 0.5722 \\
          & BERT  & 0.6144 & 0.6765 & \textbf{0.5957} \\
    \multicolumn{1}{l}{QQP} & GPT-2 & 0.7446 & 0.7361 & \textbf{0.7389} \\
          & BERT  & 0.7212 & 0.7198 & 0.7204 \\
    \bottomrule
    \end{tabular}%
    }
  \caption{Embeddings performance in different datasets (fine-tuned models in the corresponding task).}
  \label{tab:embeddingsclassifiers}%
\end{table}%

Unlike the well-formed text where language models are usually pre-trained, tweets can contain twitter-specific entities, abbreviations, and emoticons among other distinctive features. Such entities are candidates to be split into sub-word units since they are hardly covered in the corpus where language models are trained. Figure \ref{fig:tok2enlenghtdist2} shows the word length distribution in our dataset measured in number of tokens (words and word units) and compares it with two datasets used in sentence classification tasks that are part of the GLUE benchmark \cite{wang-etal-2018-glue}: the Corpus of Linguistic Acceptability (CoLA) \cite{warstadt-etal-2019-neural}, with sentences from the linguistic literature domain, and Stanford Sentiment Tree Bank (SST-2) \cite{socher-etal-2013-recursive}, which includes movie reviews.

In general, words in tweets are split in more tokens than words in the other datasets. As shown in figure \ref{fig:sub1}, BERT and GPT-2 have similar distributions for words split in less than 10 tokens, however for words with more tokens the patterns are different. Manual inspection shows that URLs are often split in many tokens. If we remove the URLs (see figure \ref{fig:sub2}) the two  distributions become very similar. In general, word length distribution in our dataset is very different compared to CoLA and SST-2, where most of the words are represented using 1 to 3 or 4 tokens. Beyond the greater number of sub-words in our dataset and the difference between BERT and GTP-2 when representing URLs we did not find further significant differences between both approaches at this level. Thus, we move on with our analysis at the embedding layer.

We evaluate the embeddings learnt by the language models after fine-tuning on the bot detection task in the same classification task. For every tweet, we concatenate the embeddings of all their tokens and feed them to a linear fully connected layer to train a softmax classifier. Due to memory restrictions we limit the number of token in each tweet to 128 maximum. 
We carry out similar experiments with CoLA and SST-2 datasets, training classifiers on their sentence classification tasks. We also include the Quora Question Pairs\footnote{\url{https://data.quora.com/First-Quora-Dataset-Release-Question-Pairs}} (QQP) and the Microsoft Research Paraphrase Corpus (MRPC) ~\cite{dolan-brockett-2005-automatically} datasets from GLUE, which require to classify pairs of sentences from social question answering and news, respectively. 

The evaluation of the embedding-based classifiers are reported in table~\ref{tab:embeddingsclassifiers}. In the bot detection task, BERT generates embeddings that help to learn better classifiers than GPT-2, which contrasts with our early findings, where GPT-2 outputs contribute to learn more accurate classifiers. Nevertheless, the embeddings alone do not seem to be a reliable indicator of the performance of the models. Fine-tuning BERT on CoLA, SST-2, QQP, and MRPC systematically produces more accurate results than GPT, according to~\cite{devlin2018BERT} and~\cite{radford2018OpenAITransformer}, and GPT-2, according to our own experimentation. However, BERT embeddings only generate better classifiers in SST-2 and MRPC, while GPT-2 embeddings are better in CoLa and QQP. Thus, we disregard the embeddings and focus our analysis on the transformer hidden states.


\section{Exploring hidden states}
\label{sec:hid}

We aim at quantify the potential contribution of each layer in the transfomer to the bot detection task. We connect the hidden state of the classification token (\textit{CLS} in case of BERT or the end token in the case of GPT-2) at each layer of the fine-tuned model to a linear layer and train a softmax classifier for the bot detection task. In this experiment, we train each classifier for 2 epochs with a learning rate of 2e-5 and batch size 8. In addition, we carry out the same experiment for language models fine-tuned on CoLA, SST-2, QQP and MRPC. 

\begin{figure*}[h!]
    \centering
    \begin{subfigure}[t]{.19\textwidth}
        \centering
        \includegraphics[width=\textwidth]{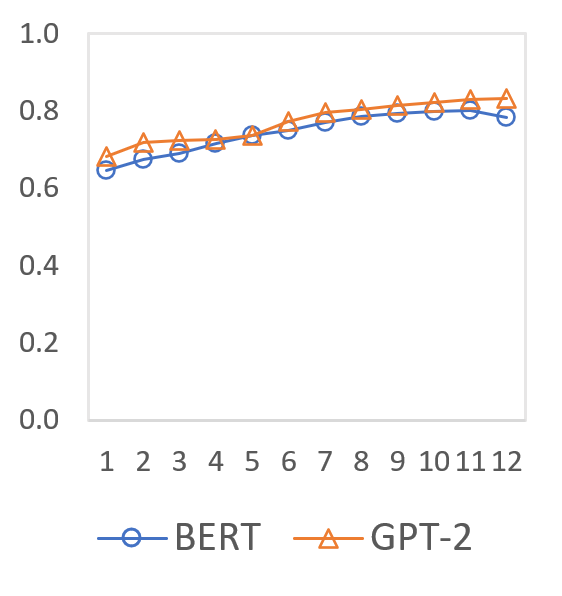}
        \caption{Bot detection}
        \label{fig2:sub1}
    \end{subfigure}%
    \begin{subfigure}[t]{.19\textwidth}
        \centering
        \includegraphics[width=\textwidth]{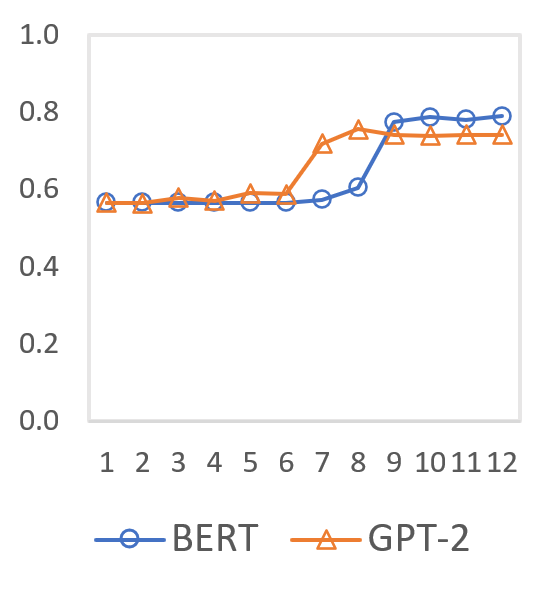}
        \caption{CoLA}
        \label{fig2:sub2}
    \end{subfigure}
    \begin{subfigure}[t]{.19\textwidth}
        \centering
        \includegraphics[width=\textwidth]{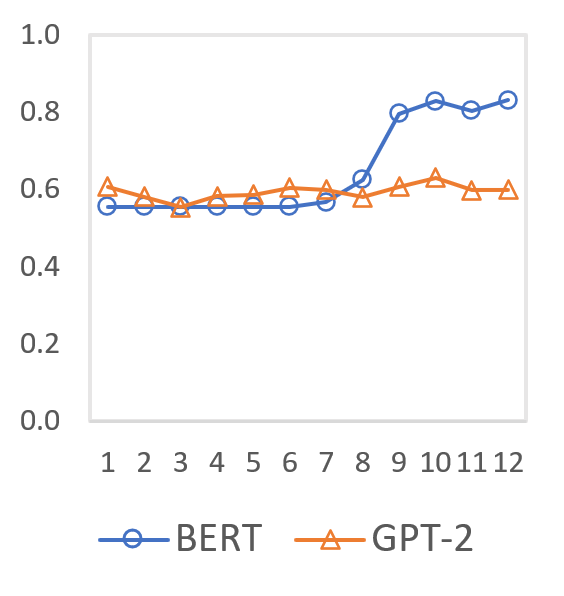}
        \caption{MRPC}
        \label{fig2:sub3}
    \end{subfigure}
    \begin{subfigure}[t]{.19\textwidth}
        \centering
        \includegraphics[width=\textwidth]{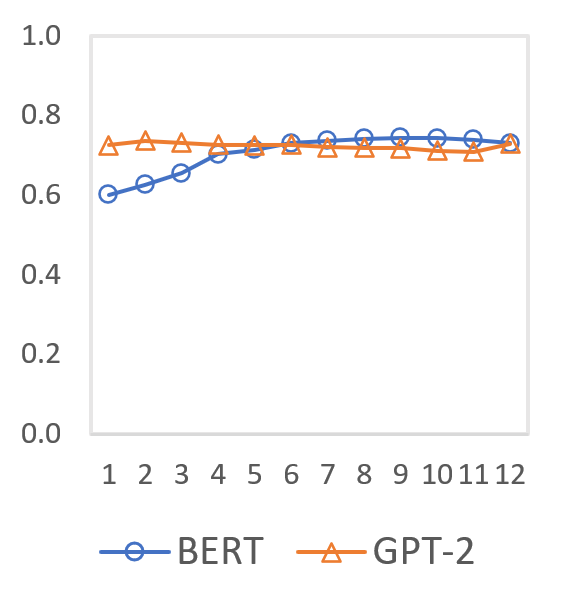}
        \caption{QQP}
        \label{fig2:sub4}
    \end{subfigure}
    \begin{subfigure}[t]{.19\textwidth}
        \centering
        \includegraphics[width=\textwidth]{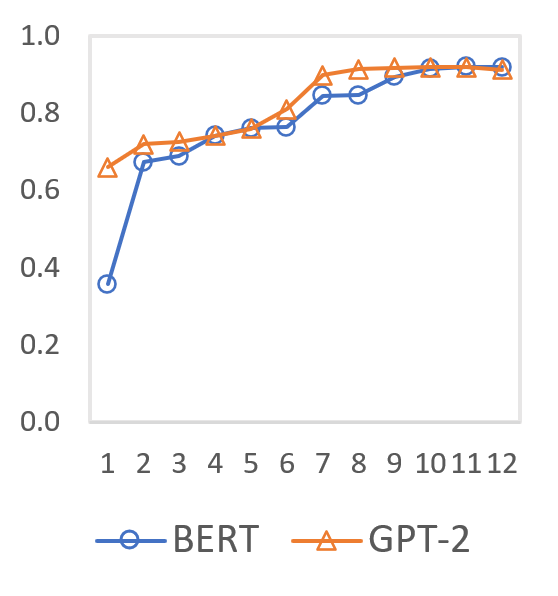}
        \caption{SST-2}
        \label{fig2:sub1}
    \end{subfigure}%
    \caption{Layer analysis of BERT and GPT-2 fine-tuned on bot detection, SST-2, CoLA, MRPC and QQP. X-axis corresponds with the hidden state layer used while  Y-axis is the f-score using that hidden state layer. }
    \label{fig:cls_layerxlayer}
  \end{figure*}

The results depicted in figure \ref{fig:cls_layerxlayer} show
that, for the bot detection task, GPT-2 hidden states generally contribute to learn slightly more accurate classifiers layer by layer than using BERT hidden states, except for layer 5. While for SST-2 the models exhibit a similar pattern, for CoLA, MRPC and QQP tasks, the pattern is different and recurrent, showing that in the first layers GPT-2 hidden states help to learn classifiers with greater or similar f-measure than BERT hidden states, and from the intermediate to the final layers BERT hidden states contribute to learn classifiers with similar or greater f-measure. The slight but constant margin in bot detection shows evidence that the internal representations of GPT-2 do a better job at encoding tweet information for this task. 



\subsection{Grammatical Analysis}\label{pos}
We investigate the ability of language model hidden states  to capture grammatical information upon fine tuning for bot detection. As grammatical tasks we use part-of-speech tagging and chunking. 
Our goal is to determine whether the hidden states of different words with the same part-of-speech or chunk tag are positioned in the same region of the distributional vector space. 

  \begin{figure*}[ht!]
    \centering
    \begin{subfigure}[t]{.45\textwidth}
        \centering
          \includegraphics[width=\textwidth]{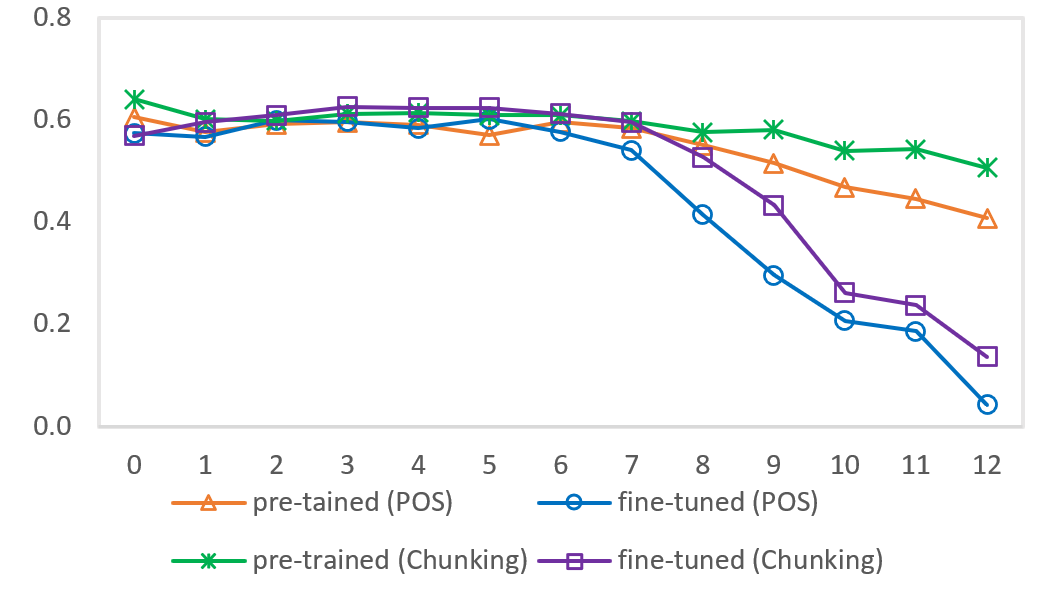}
          \caption{BERT}
          \label{fig:nmibertgrammar}
    \end{subfigure}%
    \begin{subfigure}[t]{.45\textwidth}
        \centering
        \includegraphics[width=\textwidth]{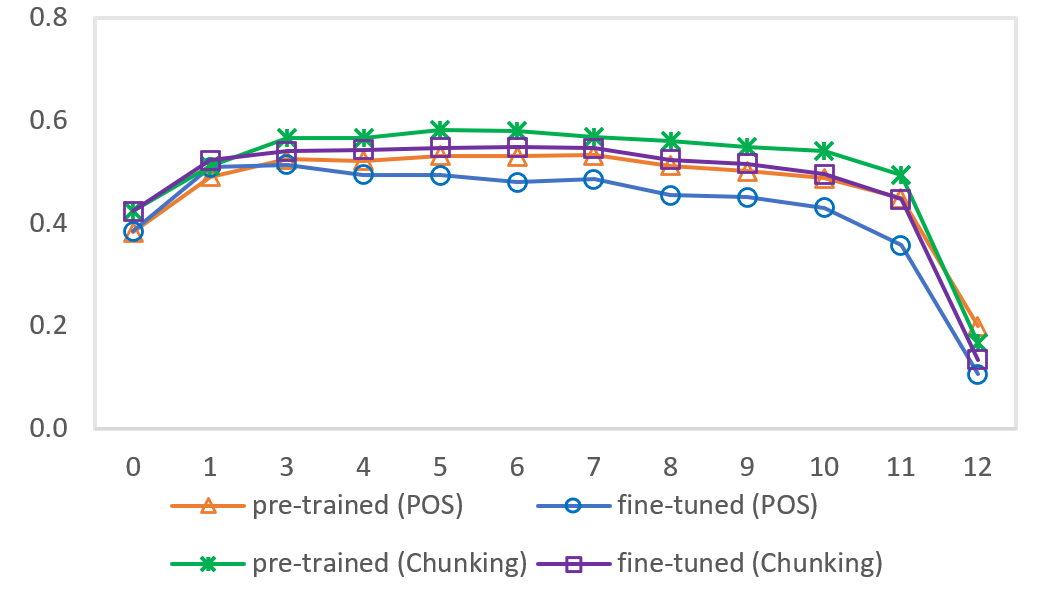}
     \caption{GPT-2}
        \label{fig:nmigpt2grammar}
    \end{subfigure}
    \caption{NMI of the hidden state clusters and part-of-speech tags and chunk labels. We compare the results obtained with pre-trained language models vs. their versions fine-tuned for  the bot detection task.}
    \label{fig:nmipos}
  \end{figure*}

For the chunking task we use regular text from CONLL-2000 dataset~\cite{CoNLL-2000}. For the part-of-speech tagging task we annotate our dataset using TweetNLP tagger~\cite{Owoputi2013ImprovedPT}. This tool reported 93.2\% accuracy, and in addition to common word categories like verbs and nouns, includes twitter-specific labels like emoticons or hashtags. Then, we cluster the hidden states and outputs for each tagged word using k-means and evaluate the clusters using normalized mutual information (NMI). This procedure was used in  ~\cite{peters-etal-2018-dissecting,jawahar-etal-2019-bert} to assess whether hidden states capture syntactic information. 


We tested pre-trained and fine-tuned models to figure out how the adjustment to the bot detection task affects the language model hidden states. We generate clusters out of the hidden states and outputs for 3000 labeled tokens randomly selected from the POS and chunking datasets. We repeat this experiment 10 times and report in figure \ref{fig:nmipos} the average NMI scores for these clusters. 

The hidden states in pre-trained BERT produce, in general, greater NMI scores than GPT-2's for the POS-tagging task and also in the chunking task. The curves, high at the beginning and decreasing towards the end, are in line with previous findings \cite{jawahar-etal-2019-bert,tenney-etal-2019-bert}, which reported that BERT encodes mainly syntactic information in its first layers, while the last layers are more task-specific.

However, when the models are fine-tuned for the bot detection task, BERT's hidden states and outputs are significantly worse at encoding POS and chunking information from layer 8 to 12, while GPT-2 holds a similar pattern as the pre-trained model for POS tagging and chunking. This suggests that, upon fine-tuning for bot detection, GPT-2 manages to preserve the grammatical information contained in its internal representations, while BERT on the other hand seems to lose part of it. 

\subsection{Twitter Entitites}
\begin{figure}[htbp!]
 \begin{minipage}{0.48\textwidth}
       \begin{subfigure}{0.47\linewidth}
         \includegraphics[width=1\linewidth]{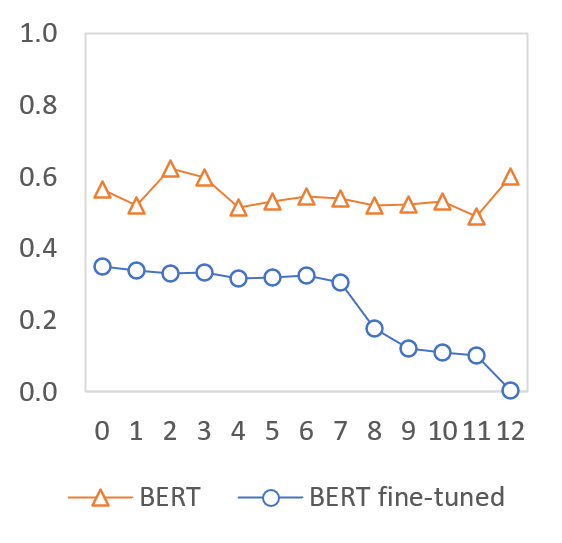}
       \end{subfigure}
       \begin{subfigure}{0.47\linewidth}
         \includegraphics[width=1\linewidth]{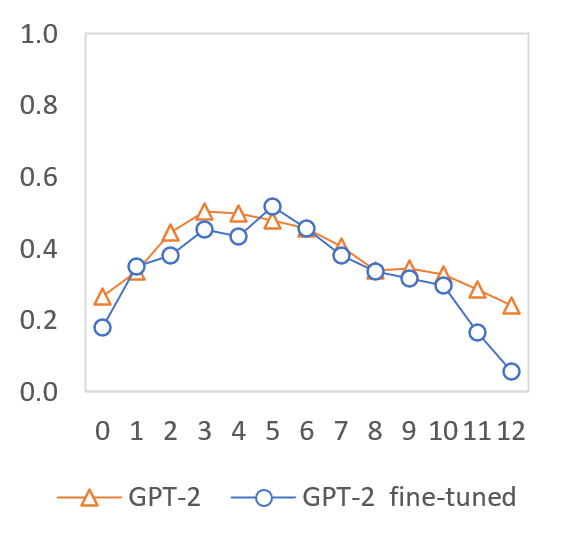}
       \end{subfigure}
       \caption{NMI evaluation of clusters out of hidden states and outputs for twitter entities.}\label{fig-1:NMI-entities}
     \end{minipage}
\end{figure}

We also look into hidden states and outputs for twitter entities  (\textit{user mention}, \textit{hashtag}, and \textit{url}) since GPT-2 learns to spot these entities with more accuracy than BERT. We use k-means to cluster the language model hidden states and outputs for 3000 entities randomly sampled from our dataset. We repeat this experiment 10 times and report in figure \ref{fig-1:NMI-entities} the average NMI of the clusters. 

Note the negative effect of fine-tuning BERT for bot detection when representing twitter entities. While pre-trained BERT encodes twitter entities both in hidden states and outputs, the fine-tuned model generates consistently lesser NMI values for these contextual representations. Also, the NMI drop is particularly acute from layers 8 to 12. After fine-tuning, hidden states and outputs of GPT-2 follow a similar pattern to the pre-trained model, encoding twitter entities mainly between layers 2 to 10. However, in the last layers GPT-2 representations for these entities are worse than in the pre-trained model. 

\subsection{Contextual Similarity}
\begin{figure*}[htbp!]
    \centering
    \begin{subfigure}[t]{.24\textwidth}
        \centering
        \includegraphics[width=\textwidth]{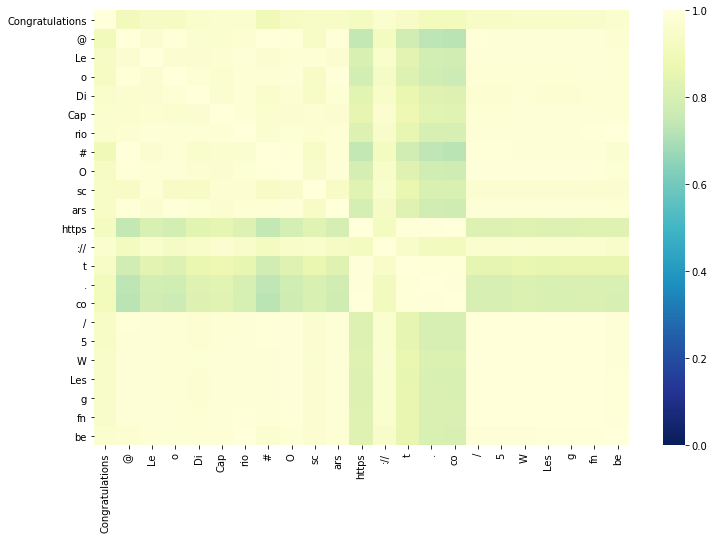}
        \caption{GPT-2 pre-trained}
        \label{fig-context:sub3}
    \end{subfigure}
    \begin{subfigure}[t]{.24\textwidth}
        \centering
        \includegraphics[width=\textwidth]{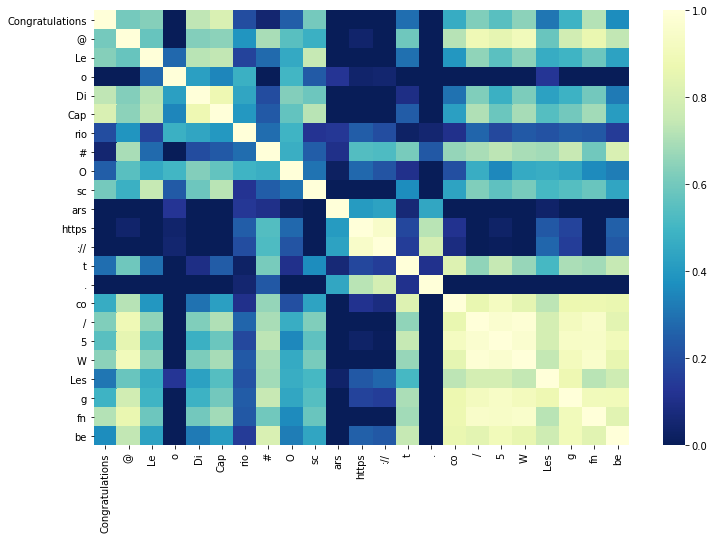}
        \caption{GPT-2 fine-tuned}
        \label{fig-context:sub4}
    \end{subfigure}
    \begin{subfigure}[t]{.24\textwidth}
        \centering
        \includegraphics[width=\textwidth]{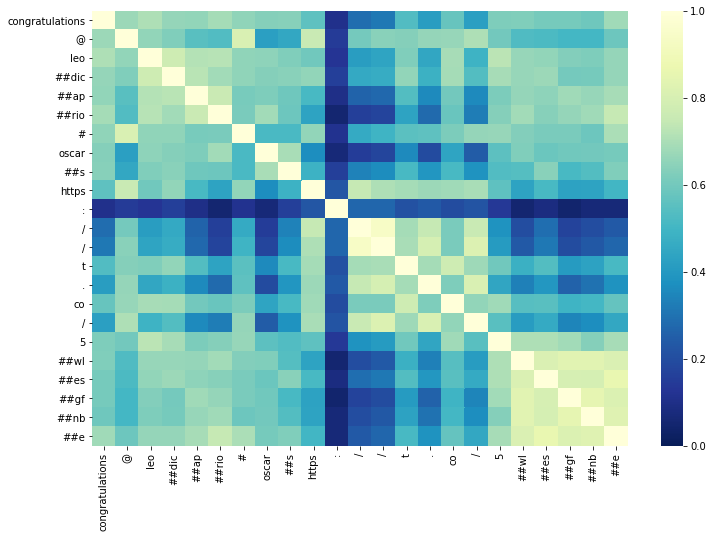}
        \caption{BERT pre-trained}
        \label{fig-context:sub1}
    \end{subfigure}%
    \begin{subfigure}[t]{.24\textwidth}
        \centering
        \includegraphics[width=\textwidth]{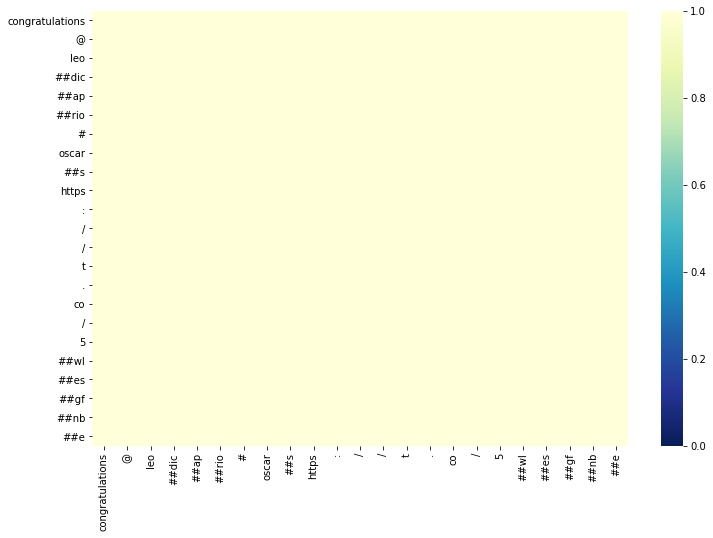}
        \caption{BERT fine-tuned}
        \label{fig-context:sub2}
    \end{subfigure}
    \caption{Contextual similarity  of all words and sub-word units pairs in a tweet computed on the language model output representations of the pre-trained and fine-tuned models. }
    \label{fig:contextualsimilarity}

  \end{figure*}

To gain additional insight, we investigate the contextual embeddings generated by each model and how these outputs change from pre-trained to fine-tuned models. Let us focus \textit{e.g.}, on the tweet: \textit{"Congratulations @LeoDiCaprio \#Oscars https://t.co/5WLesgfnbe"}. Cosine similarity values of the outputs for tokens in this tweet are shown in figure \ref{fig:contextualsimilarity}, where lighter colors indicate greater similarity and darker means the contrary. 

Note that pre-trained GPT-2 outputs are very similar, except some of the URL tokens (\textit{https}, \textit{t}, $\cdot$, and \textit{co}) which are more similar among them. However, in the fine-tuned model the outputs become less similar and some groups of similar outputs emerge that are roughly associated with entities, such as \textit{user mentions}, \textit{hashtags}, and \textit{URLs}. On the other hand, BERT follows the opposite pattern. That is, the pre-trained model generates similar outputs for entities (\textit{leo \#dic \#ap \#rio},  and \textit{oscar \#s}) and URL substrings. However, when the model is fine-tuned all the outputs are very similar among them and no segmentation is noticeable in the vector space.  

\begin{figure}[htbp!]
  \begin{minipage}{0.48\textwidth}
       \begin{subfigure}{0.47\linewidth}
          \includegraphics[width=1\linewidth]{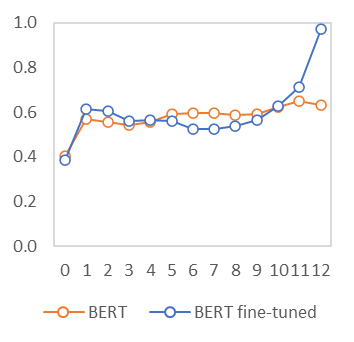}
       \end{subfigure}
       \begin{subfigure}{0.47\linewidth}
         \includegraphics[width=01\linewidth]{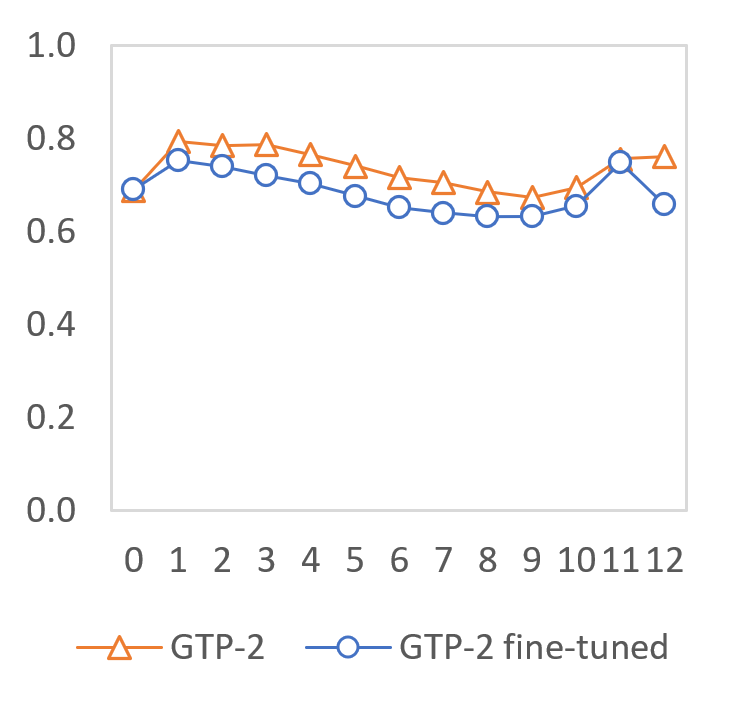}
       \end{subfigure}
       \caption{Average intra-sentence contextual similarity.}\label{fig-1:avgintrasentence}
     \end{minipage}
\end{figure}

To investigate whether this observation generalizes we randomly sample 100,000 tweets from our dataset and calculate the average cosine similarity between the outputs of the tokens in each tweet. The results in figure~\ref{fig-1:avgintrasentence}  confirm that pre-trained BERT outputs are less similar, and hence more distributed in the vector space than pre-trained GPT-2 outputs. These results are aligned with \newcite{ethayarajh2019contextual} findings for text extracted from the SemEval Semantic Textual Similarity task, where GPT-2 outputs for samples of word pairs were more similar than BERT's. Fine-tuned for bot detection, GPT-2 outputs become less similar, while BERT outputs turn into very similar. 

\subsection{Discussion}\label{discussion}
Fine-tuning on the bot detection task modifies the hidden states and outputs of both BERT and GPT-2, although in different ways. pre-trained BERT encodes grammatical information through all layers regardless whether it was tested on regular text as in the CONLL-2000 dataset or on tweets as in the part-of-speech experiments (figure \ref{fig:nmibertgrammar}). Nevertheless, fine-tuned BERT encoding of grammatical information is poorer in the final layers. After fine-tuning, the final layers change the most \cite{kovaleva-etal-2019-revealing} to adjust the representation to the corresponding task, and hence grammatical information vanishes in these layers. 


BERT encoding of twitter entities (figure \ref{fig-1:NMI-entities}) exhibits a similar pattern to the grammatical information (figure \ref{fig:nmibertgrammar}). However, in this case there is larger gap between the encoding of entities in the pre-trained and the fine-tuned model across all layers and not limited to the last layers as in the grammatical information analysis. We think that such a change across all layers indicates that BERT modifies to a great extent the representation of twitter entities since they are relevant for the bot detection task. 

According to~\newcite{kovaleva-etal-2019-revealing}, \textit{CLS} takes most of the focus of BERT's self-attention during fine-tuning. Therefore, we also hypothesize that the changes in the BERT contextual embeddings and representations associated with twitter entities are related to the increased attention on \textit{CLS}. This would contribute to explain the loss of linguistic information and the high similarity of the outputs upon fine-tuning (figure \ref{fig-1:avgintrasentence}).   

On the other hand, GPT-2 shows similar patterns in the pre-trained and fine-tuned models when encoding grammatical information (figure \ref{fig:nmigpt2grammar}) and twitter entities (figure \ref{fig-1:NMI-entities}). Similarly the contextual representations are very stable upon fine-tuning (figure \ref{fig-1:avgintrasentence}). We think this is a consequence of a) the left-to-right text processing strategy followed in GPT-2 and b) the use of the last token's language model output for classification during fine-tuning. The model needs to process the whole tweet token by token in order to be able to adjust the last token representation for the bot detection task. An analysis of the self-attention mechanism in GPT-2 could identify on which tokens the model is focused during each fine-tuning step. Nevertheless such analysis is out of the scope of this research, where we focused on hidden states and outputs.

\section{Conclusions}\label{conclusion}
In the bot detection task the generative transformers GPT and GPT-2 were more accurate than the bidirectional masked language model learnt by BERT. We show that this behaviour is also true when GTP-2 and BERT are fine-tuned to detect entities that are twitter-specific such as user mentions, hashtags and URL. In our dataset bots and humans show different usages patterns for these entities.  

Since BERT generally reports higher accuracy for the text classification tasks in GLUE, we investigate both language models in order to understand why in the bot detection task the generative transformers were more accurate. At the vocabulary level we found that words in our dataset are split into more subwords than words found on regular text datasets, and when removing URL the word lenght distribution turn into very similar. Next, we discarded the word embeddings as a reliable indicator of the model performance in text classification tasks. 

By probing the hidden states of both models in the bot detection task we found that the hidden states of GPT-2 contributed to learn more accurate classifiers than BERT's, across all layers. This was a distinct pattern compared to the models fine-tuned for some text classification tasks in GLUE (CoLA, QQP, and MRPC). 

We show that after fine-tuning BERT hidden states encode less grammatical information than the pre-trained model particularly in the final layers, and the encoding of twitter entities is affected across all layers. In addition, while pre-trained BERT contextualized embeddings are more distributed, after fine-tuning they become more similar. 


On the contrary, pre-trained and fine-tuned GPT-2 do not show acute changes about  the encoding of grammatical information in the hidden states and outputs. In the twitter entities spotting probe GPT-2 hidden states change but only on the last layers, and the contextual representations were more distributed after fine-tuning which indicates that the model was discriminating more the representations. We think that the stability of the model after fine-tuning is due to the use of the last token output in fine-tuning and the left-to-right text processing used in GPT-2 which enforce to process the whole tweet before adjusting the last token representation for the bot detection task.


Future work includes the extension of this study to other tasks based on Twitter data to assess whether the language model learnt by GPT-2 is more suitable to process twitter data in general beyond the bot detection task. In addition, we would like to investigate the role of the next sentence prediction learning objective used in BERT since twitter messages are short and typically consist of only one long sentence. 

\section*{Acknowledgements}
This research was funded by the EU Horizon 2020 grant European Language Grid (ELG-825627).

\bibliography{anthology,custom}
\bibliographystyle{acl_natbib}

\end{document}